# SYMMETRY AND VARIANCE. GENERATIVE PARAMETRIC MODELLING OF HISTORICAL BRICK WALL PATTERNS.

## SEVGI ALTUN, MUSTAFA CEM GÜNEŞ, YUSUF H. SAHIN, ALICAN MERTAN, GOZDE UNAL and MINE ÖZKAR

Name: Sevgi Altun (b. 1995)
Profession: Graduate Student, Research Assistant
Fields of interest: Design Computation
Affiliation: İstanbul Technical University
E-mail: altunse19@itu.edu.tr

Name: Mustafa Cem Güneş (b. 1993)
Profession: Graduate Student
Fields of interest: Design Computation
Affiliation : İstanbul Technical University
E-mail:gunesmus@itu.edu.tr

Name: Yusuf H. Sahin (b. 1992)
Profession: Graduate Student, Research Assistant
Fields of interest: AI & Data Engineering
Affiliation: İstanbul Technical University
E-mail: sahinyu@itu.edu.tr

Name: Alican Mertan (b. 1994)
Profession: Graduate Student
Fields of interest: AI & Data Engineering
Affiliation: University of Vermont
E-mail: alican.mertan@uvm.edu

Name: Gözde Ünal (b. 1974)
Profession: Professor
Fields of interest: AI & Data Engineering
Affiliation: İstanbul Technical University
E-mail: gozde.unal@itu.edu.tr

Name: Mine Özkar (b. 1976)
Profession: Professor
Fields of interest: Design Computation
Affiliation: İstanbul Technical University
E-mail: ozkar@itu.edu.tr

***Abstract:*** *This study integrates artificial intelligence and computational design tools to extract information from architectural heritage. Photogrammetry-based point cloud models of brick walls from the Anatolian Seljuk period are analysed in terms of the interrelated units of construction, simultaneously considering both the inherent symmetries and irregularities. The real-world data is used as input for acquiring the stochastic parameters of spatial relations and a set of parametric shape rules to recreate designs of existing and hypothetical brick walls within the style. The motivation is to be able to generate large data sets for machine learning of the style and to devise procedures for robotic production of such designs with repetitive units.*

## INTRODUCTION

Machine learning applications in architectural heritage comprise tasks of identifying patterns and components of a building in order to accurately and rapidly create building information models from survey data (Grilli and Remondino, 2020, p. 379). One caveat for these tasks is the need for large training data sets that do not exist but have to be synthetically generated based on documented real examples. Independently, details of construction are important to capture and depict in building information models of historical structures for a holistic understanding of the heritage as well as its reconstruction. In architectural heritage, construction and ageing irregularities impact the legibility of patterns. We offer a methodology to model historical brick patterns simultaneously, considering both the inherent symmetries and irregularities. Working in 2D in accordance with the exposed wall surface, we process the geometric organisation of bricklaying in Anatolian Seljuk heritage to represent its interrelated design and construction and offer a generative model to be used for creating myriads of existing and possible designs of the style. We devise a parametric shape grammar, i.e. a rule-based definition of bricklaying relying on the symmetries of the units and of the pattern. Our reference data is photogrammetry-based point cloud models in which bricks of a historical wall are represented as unique rectangular shapes. We transfer the features and their variance statistics into parametric shape rules, which are sequentially applied to recreate a wall. Labels are used in the parametric shape rules to disrupt the symmetry group of any rectangular brick for Euclidean transformations. Although symmetry is a key feature and the pattern remains the same, each brick is a unique variant.

## METHODOLOGY

While interpreting historical brick walls as repeating rectangular units laid with symmetry operations, we uniquely capture variance in each unit based on the photogrammetric data of real-world examples. Our initial data comes from an interior brick wall from a 13th century building known as Sahip Ata Hanikah in Konya (Figure 1). We develop our model and test its implementation with other brick walls from the same era and region which show particular characteristics and a degree of wear. Based on the data from the point cloud models as input for the parameters, we devise parametric shape rules to serve the reproduction of the brick wall with horizontal stacking. Although there have been studies that focus on bricklaying in parametric models (Afsari *et al*., 2014, pp. 49–64; Imbern, 2014, pp. 211-220), they are neither based on heritage nor do they use real-world data. And when parametric modelling is used in architectural heritage reconstruction, parameters are not directly extracted from actual survey data (Garg and Das, 2013, pp. 105–134; Li *et al*., 2013, pp. 697-703; McLennan and Brown, 2021, pp. 4040-4055). Limited research has implemented shape grammars to define parametric models for various bricklaying patterns and surface qualities (Vazquez *et al*., 2020, p. 262-274). The computational aspect of parametric shape rules is one of the challenges of shape grammar applications. Studies focus on interpreters applying parametric rules (Grasl and Economou, 2018, pp. 208-224; Kramer and Akleman, 2019; Nasri and Benslimane, 2017, pp. 1-20; Yue *et al.*, 2009, pp. 757- 770). The researchers mainly describe shapes through graphs. There are limited examples of the application of parametric rules for design purposes. Oberhauser *et al*. (2015, pp. 21-39) used parametric shape grammars to create different variations of an aircraft design using the parameters and constraints of prior design samples. The study hereby presented uses the shape grammar interpreter Sortal GI (Stouffs, 2018, pp. 453-462) to define and apply parametric shape rules of bricklaying to recreate the existing brick wall. The application of the proposed shape rules on the interpreter quickly generates multiple examples of brick walls made up of the same units and unit relations with the case study.

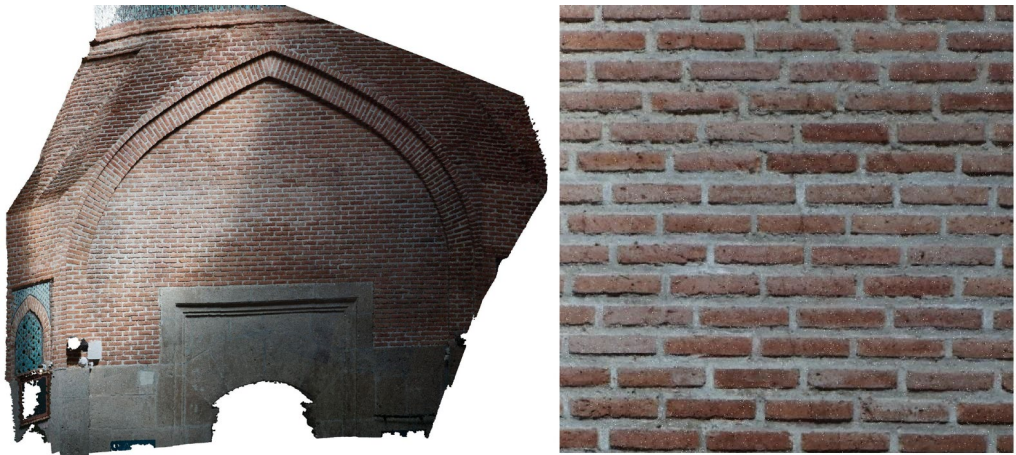

*Figure 1* Example images from photogrammetric models.

Photogrammetry offers rapid, cost-efficient, and reasonably accurate documentation of architectural heritage (Moyano *et al*., 2020, pp. 303-314). The process delivers a point cloud that stores colour and coordinate data that can be modelled into a look alike of the real artefact. The data of the selected Seljuk era brick wall visually displays rows of bricks and the binding mortar. However, this data lacks any semantic information about the bricklay patterns and its separate components. Our work pays due attention to the features that make up the wall in order to complete its detailed model. For instance, the mortar binds the bricks together and fills in the space in between them. A brick has a twofold symmetry on its exposed face. On the wall surface, bricks are placed repetitively with translation and glide reflection in rows creating various patterns.

We use photogrammetry-based point clouds with reduced density to minimise the calculation cost (Figure 2). Applying semantic segmentation to label the points as either brick or mortar, we import the model into the Grasshopper environment in Rhinoceros 3D to define and compute the information of the brick units. Since our focus is on the flat brick walls with horizontal stacking, other architectural elements and the curved parts in the data are manually excluded. We characterize the bricks as vertically aligned rectangles and calculate the dimensions and distances between adjacent rectangles.

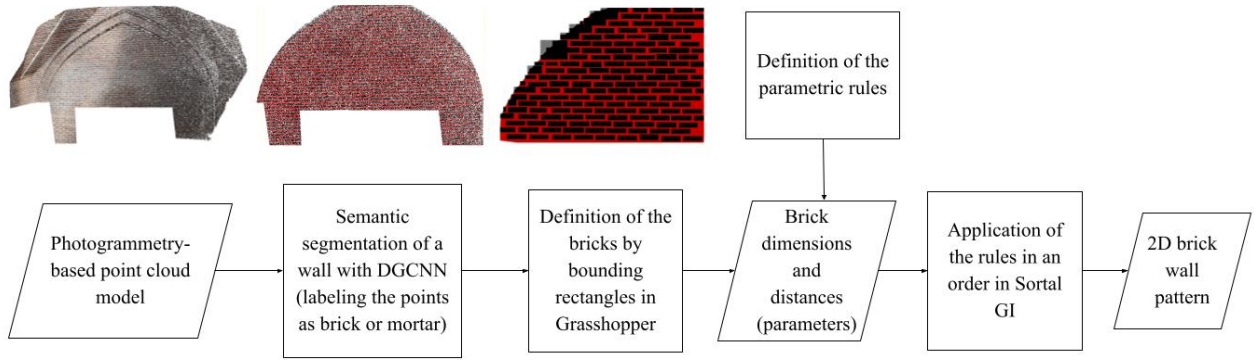

*Figure 2* The flow of data process

## PREPARATION OF THE DATA

The point cloud data obtained through photogrammetry is classified as brick or mortar for each point using semantic segmentation with a deep neural network on the data we acquire from historical sites. Considering the advances of the deep neural networks recently (Qi *et al*., 2017, p.30; Zhang *et al*., 2019, pp. 1607-1616; Sahin *et al*., 2022, pp. 610-618), and their success in architectural heritage (Matrone *et al*., 2020, pp. 1419-1426), we have chosen to prepare the data to be post processed by training a Dynamic Graph Neural Network (Wang *et al*., 2019, pp. 1-12). We apply a post-processing algorithm for getting each brick point cluster individually. Subsequently, we fit rectangular shapes around the boundaries of the brick points based on their proximity to each other to simplify the problem and follow the relations of the brick units (Figure 3(a)). Consequently, in the Grasshopper environment Rhinoceros, we segregate the edges of the brick blocks and calculate the distances of the edges as parameters of relations between units. Instead of using uniform random distribution in our parametric rule definitions, we use statistical information from the real samples.

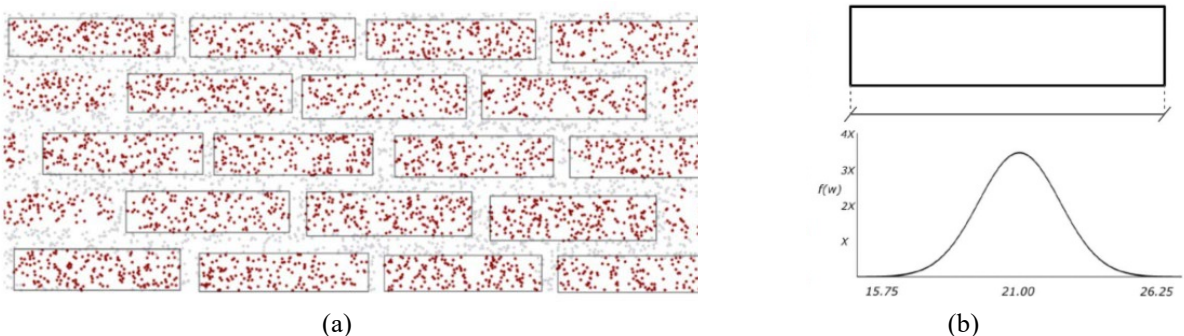

(a)                                                                            (b)

*Figure 3* (a) The brick clusters as rectangles., (b) A Gaussian distribution for the width of a brick.

## PARAMETRIC RULES FOR BRICKLAYING

The selected basic brick wall is composed of naked whole bricks with horizontal stacking. A visual analysis of it shows separate sets of parameters for the bricks and for the relations between them. The primary parameters for the bricks are the width and the height, while the parameters for the relation are distances between the edges of the shape on the left side of the rule and the shape added on the right side (Figure 4). The depth of the bricks is omitted here since the point cloud model lacks the information about the actual depth of the bricks. The only data it carries regarding the depth is the level difference between the exposed surface of a brick and the surface of the mortar. With these parameters, we define the lowest and highest values for the brick dimensions and distances between the bricks which we express as Gaussian distribution (Figure 3(b)).

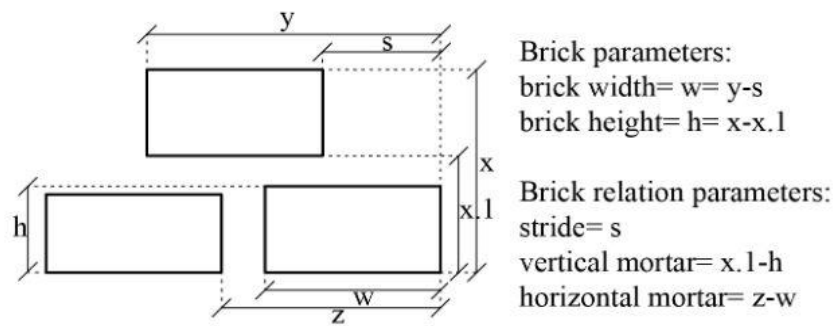

*Figure 4* Parameters of bricklaying.

We reproduce the bricklaying process with seven shape rules, five to place the bricks and two to label them to work around their axial symmetry. The labels and tags for each side of the bricks define the rules parametrically (Figure 5). The first label rule gives a direction to the brick. Even though the shape is visually symmetrical the information it carries is asymmetrical. The second label rule reflects that asymmetrical labelling along a vertical axis through its centre.

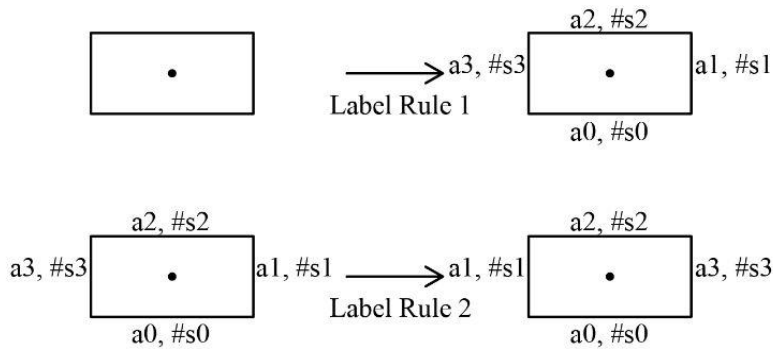

*Figure 5* The label rules.

Brick placement rules (Figure 6) are divided into three in themselves. The first set of rules is for the horizontal translation of the brick, there are two for the left and right directions. The interpreter re-quires a direction vector to define distance directives between the target and the reference tags. The distances between tagged shapes are calculated numerically with these vectors. Thus, the rule set requires rules for both directions. The rules in the second set translate and scale the brick in the horizontal direction for both edges of the wall. The rules in the third set combine two translations of the brick, horizontal and vertical, enabling the switch to a new row at the edge of the wall. We use tuple function to generate a random number, then use it as an index number to select an element from the list of extracted values. The parameters are selected from pre-defined random distance values.

We can recreate the wall using the parameters and applying the rules sequentially (Figure 7). Although this is not the exact wall from the real-world data, it is a realistic version that is procedurally modelled. The procedure can be replicated to yield any other similar instance from within the large set of walls akin to the real artefact. The significance of this simple procedure is the establishment of both local and global relationships of small and similar parts of a pattern. By holding the key to manipulating local relations as well as the global, we can govern both the whole and the parts when and if we want to modify them individually. The individual brick is a part of the global symmetric order of the brick lay for the wall while within itself may show variance in size and level as displayed in the real-world data

**CONCLUSION**

As part of a broader objective for incorporating construction and material knowledge to semi-automated information models of architectural heritage, this work models the details of a particular historical bricklaying style based on symmetry operations and parameters from the point cloud data of a real wall. Interpreting units as-is instead of idealising their features is novel and allows a more accurate transfer of the knowledge of design materialisation to an information model. The model captures the global symmetries between the parts of a repeat pattern while disrupting local ones. Next steps are to include more brick wall patterns of the period, other types of brick, the vertical bond, and to incorporate depth as well as three-dimensional geometries, deformed or curved brick wall surfaces. Our method is likely applicable to similar problems where parts in recursive relations vary in details. Since the workflow includes the definition and recording of units with their relations sequentially, our model has the capacity to be improved for robotic fabrication applications.

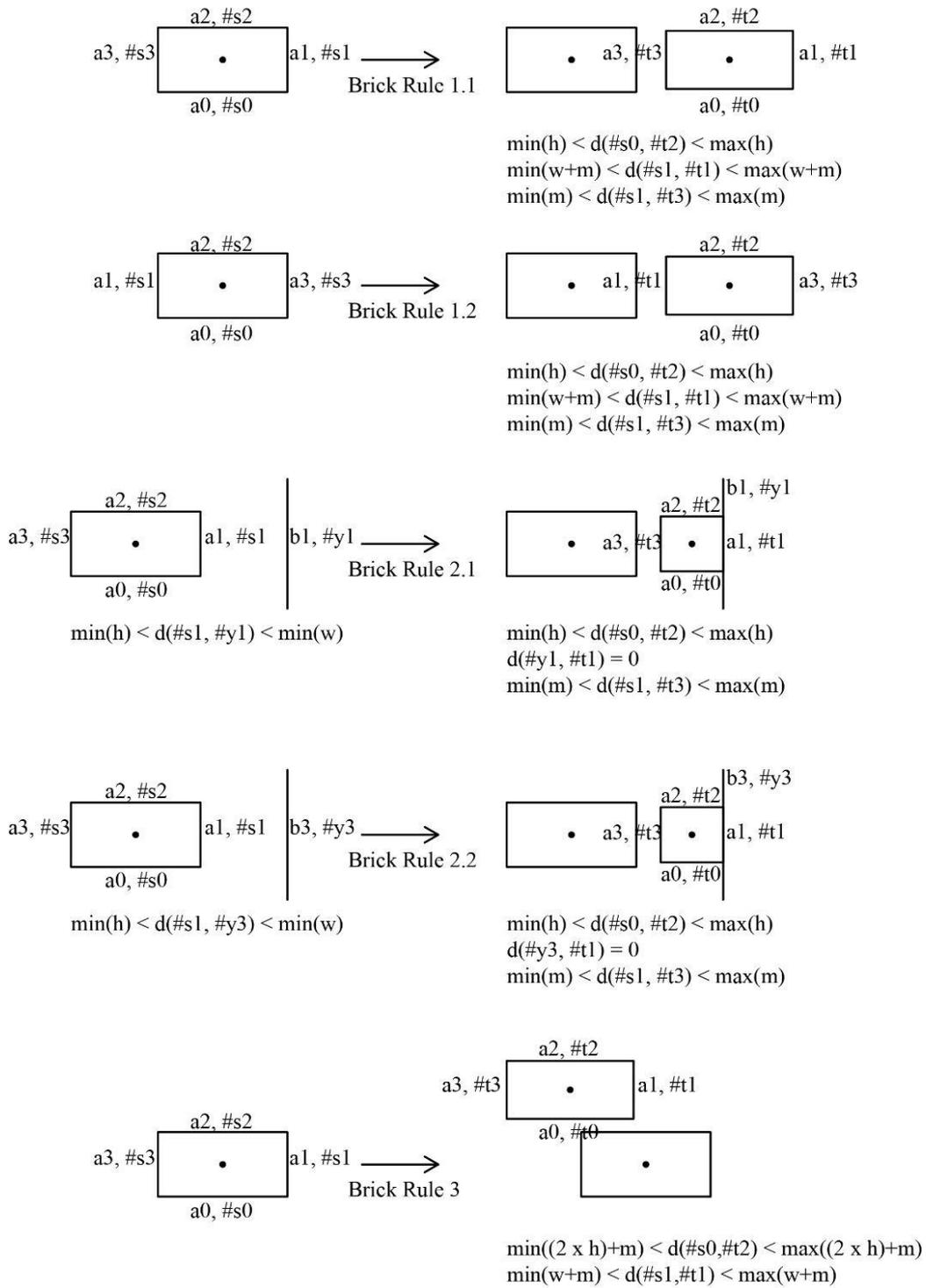

*Figure 6* The brick placement rule.

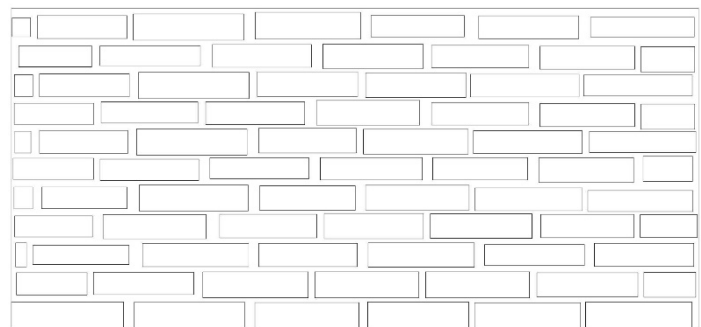

*Figure 7* Sample image from the outputs.

## ACKNOWLEDGMENT

This research is supported by TÜBİTAK (The Scientific and Technological Research Council of Turkey) under the project number 119K896.

**Sevgi ALTUN**

Sevgi Altun is an architect with a background in electrical - electronics engineering. She is a research assistant and a graduate student in the Architectural Design Computing program at Istanbul Technical University. Her research focuses on the correlation of the material construction of brick structures with design geometry.

**Mustafa Cem GÜNEŞ**

Cem Güneş is an architect with a background in computer science and a graduate student in the Architectural Design Computing program at Istanbul Technical University. He is currently working on data enrichment for deep learning in documentation tasks and semi-automatic reconstruction of historical architectural elements using visual programming tools.

**Yusuf Hüseyin ŞAHIN**

Yusuf H. Sahin is a research assistant and PhD candidate in Istanbul Technical University, Computer Engineering. He is currently working on 3D vision and point cloud analysis.

**Alican MERTAN**

Alican Mertan earned his BSc and MSc in engineering from Istanbul Technical University. He is currently a PhD at University of Vermont. He is interested in topics in artificial.

**Gözde ÜNAL**

Gozde Unal received her PhD in ECE with a minor in from North Carolina State, NC, USA. She worked as a research scientist at Siemens Corporate Research, Princeton, NJ, USA, then held positions of professor and associate professor at Sabancı . Dr. Unal joined Istanbul Technical, Department of Engineering in 2015. Currently, she is one of the founding professors of the undergraduate department of Artificial and Engineering at ITU. Dr. Unal acted as the founding director of the ITU AI Research Centre between 2018-2019. Her research interests are in representation learning, deep learning and vision.

**Mine ÖZKAR**

Mine Özkar is a Professor of Architecture at Istanbul Technical University. Her research is in visual computation, shape grammars and more recently architectural heritage in design computation. Teaming with art historians and computer scientists, she has worked on extracting visual languages of the making of geometric ornaments in medieval Anatolian architecture. Currently her collaborations implement cutting-edge machine learning methods on digital data of architectural heritage to provide accessible and easy-to-use methods to process, interpret and translate it to comprehensive models for comparative analyses.